# Hybrid Approach to Face Recognition System using Principle component and Independent component with score based fusion process

*Trupti M. Kodinariya[1], Dr. Prashant R. Makwana[2]*

*[1]Research Scholar in JJT University, Jhunjhunu, Rajasthan, India*

*Department of Computer Engineering,*

[1]trupti.kodinariya@gmail.com

*[2]Director - GRMECT Research Center*
*Rajkot, Gujarat, India*
[2]prmak1@hotmail.com

## ABSTRACT

Hybrid approach has a special status among Face Recognition Systems as they combine different recognition approaches in an either serial or parallel to overcome the shortcomings of individual methods. This paper explores the area of Hybrid Face Recognition using score based strategy as a combiner/fusion process.

In proposed approach, the recognition system operates in two modes: training and classification. Training mode involves normalization of the face images (training set), extracting appropriate features using Principle Component Analysis (PCA) and Independent Component Analysis (ICA). The extracted features are then trained in parallel using Back-propagation neural networks (BPNNs) to partition the feature space in to different face classes. In classification mode, the trained PCA BPNN and ICA BPNN are fed with new face image(s). The score based strategy which works as a combiner is applied to the results of both PCA BPNN and ICA BPNN to classify given new face image(s) according to face classes obtained during the training mode. The proposed approach has been tested on ORL and other face databases; the experimented results show that the proposed system has higher accuracy than face recognition systems using single feature extractor.

**KEYWORD**: *Face Recognition, Feature Extractor, Hybrid System, ICA, PCA, Neural Network, Score based strategy.*

## 1. INTRODUCTION

In the field of pattern recognition, computer vision has emerged as an independent field of research activities. Within this field face recognition systems have especially turned out to be the focus of interest for a wide field of demanding applications in areas such as security systems or indexing of large multimedia databases.

Holistic matching and Feature-based matching approaches are the two major classes of face recognition methods. Holistic matching is based on information theory concepts; seeks a computational model that best describes a face, by extracting the most relevant information contained in that face. Feature-based matching is based on the extraction of the properties of individual organs located on a face such as eyes, nose and mouth, as well as their relationships with each other. Some well-known example of Holistic matching and Feature-based matching are PCA[2], Fisherfaces[3], ICA[4], Neural networks[5] and Dynamic deformable template mlatching[6], Hidden Marcov Model[7] respectively.

Current Face Recognition methods, which are based on two-dimension view of face images, can obtain a good performance under constraint environment. However, in the real application face appearance varies due to different illumination, pose, and expression. Each individual classifier based on different appearance of face image has different sensitivity to these variations, which motivate to move towards hybrid approach i.e. combination of different face recognition techniques, to improve accuracy. In this paper, I focus on hybrid approach, which involves PCA and ICA system.

A typical face recognition system includes the following steps: (1) extract human face area from images, i.e. detect and locate face; (2) find a suitable representation of the face region, i.e. feature extraction; and (3) classify the representations. It is assumed that human face has been extracted from images using methods mention in[1]. In this paper, I focus on only steps 2 & 3.

## 2. A PROPOSED SYSTEM

The conventional face recognition system uses one feature domain and one classifier. Usually neural network are used as classifier therefore this conventional method named Single Feature Neural Network (SFNN) face recognition as shown in Fig. 1.

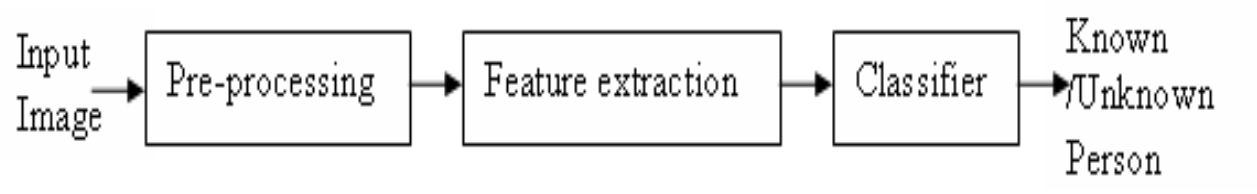

**Fig. 1:** SFNN Face recognition system

The Proposed face recognition shown in Fig.2 contains four phases.

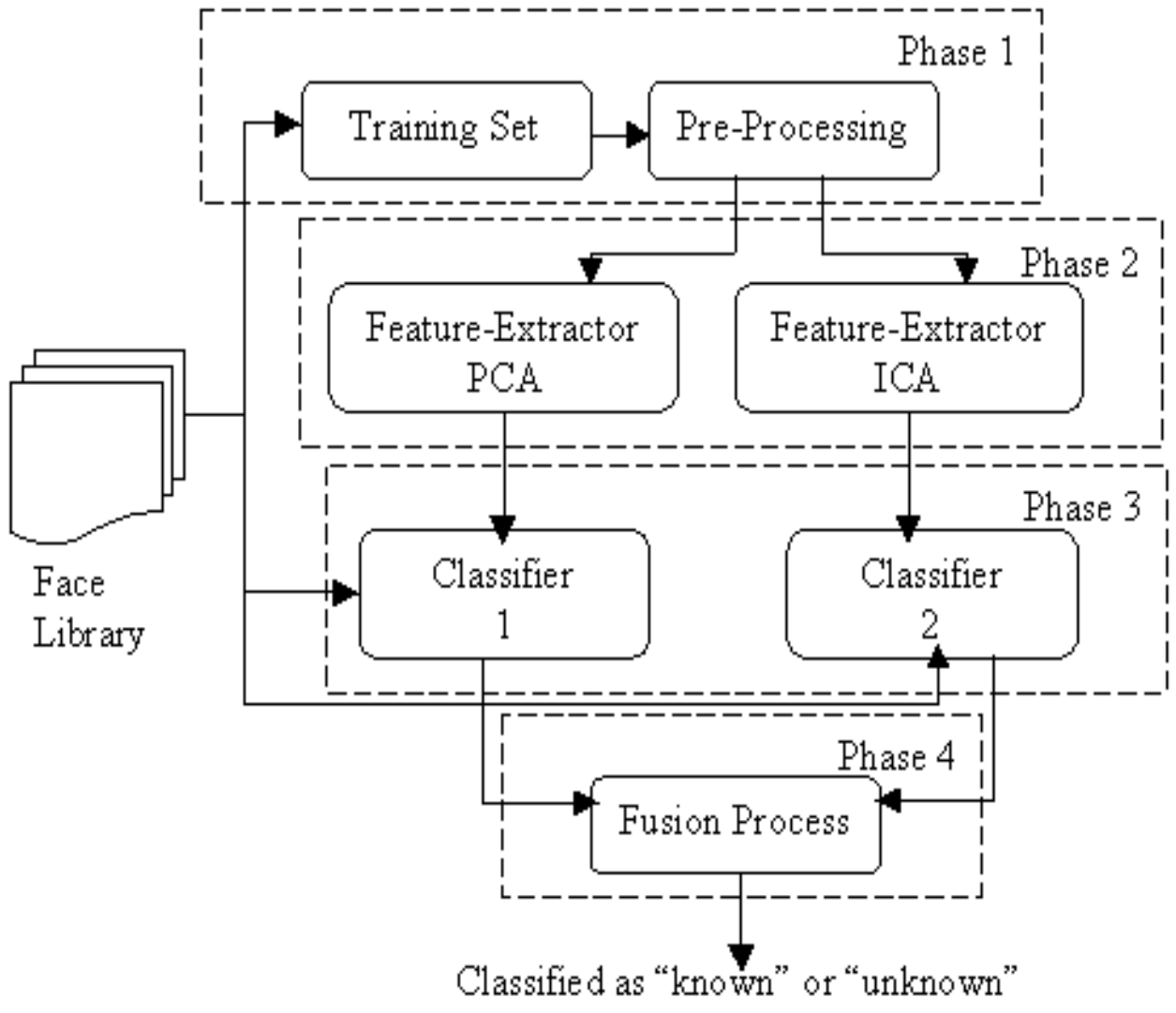

**Fig. 2:** Hybrid Multi-Feature Face recognition system

In the *first phase*, face images have been normalized by applying different normalization methods mention in [8]. In the *second phase*, different features have been extracted in parallel form normalized face images. These features are obtained from the different domains. In the *third phase*, classification has been performed which classified a given input face image(s), based on the chosen features, into one of the possibilities. This is done for each feature domain in parallel as shown in Fig. (2). finally in the *last stage*, I combined the outputs of each neural network classifiers to construct the class label of given input image(s) using score based decision strategy.

### 2.1. Pre-Processing

The face databases consider in proposed system are Olivetti research laboratory (ORL) face database and Shimon Edelman database which contain frontal view face images which already normalized to some extents, so I only perform three steps: (1) resize face images; (2) adjusting contrast (Histogram Equalization); (3) adjusting brightness (Gamma Correction).

## 2.2. Feature Extraction

Feature extraction helps to derive a meaningful representation of images by mapping high dimensional input space into a lower dimensional feature space. Not only it reduces the dimension to get a classifier that runs faster and uses less memory, but it may also improve the classification in revealing the intrinsic dimension of the observed pattern.

To design a system with low to moderate complexity the feature vectors should contain the most pertinent information about the face to be recognized. Face recognition system should be capable of recognizing unpredictability of face appearance and changing environment. The hybrid systems can have N different feature domains extracted from the normalized face images. Therefore this approach can extract more characteristics of face images for classification purpose. In this paper, I set N=2.

### 2.2.1. Principal Component Analysis

PCA is based on an information theory approach that decomposes face images into small set of feature images called "*Eigenfaces*" which may be thought of as a principle component analysis of original training set of face images.

In order to decompose, I have to extract relevant information from face image. A simple approach is to capture the variations from a collection of training face images, independent of any judgment of features (i.e. second-order statistics of the data are de-correlated) and use this information to encode and compare individuals as shown in Fig. 3a and 3b.

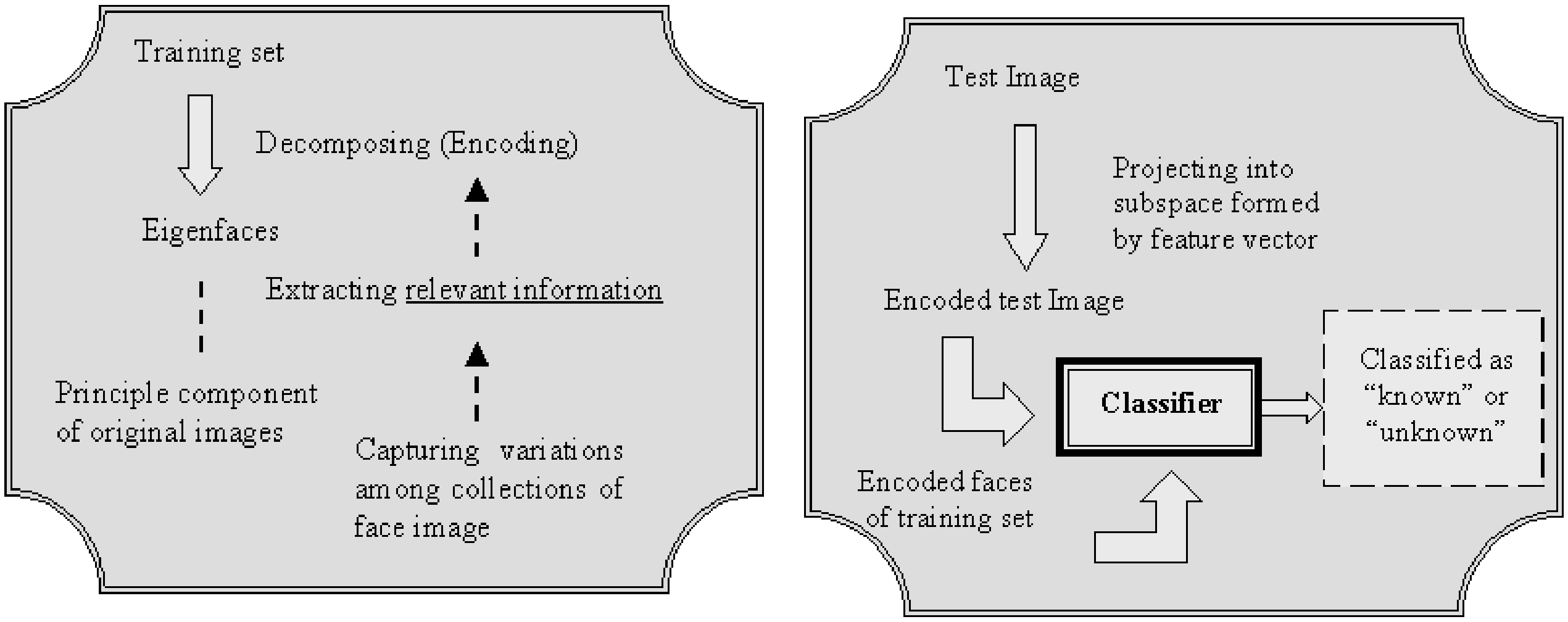

**Fig. 3a) :** Decomposing the training face images ; **3b)** Classification of the input face image

Classification is performed by projecting a input face image into subspace known as a face space spanned by "Eigenfaces" and then classifying the input face image by comparing its position in a face space with position of known individuals as shown in Fig. 3b.

The algorithm is as follows:

***Training Set***

*Step 1: Establishes the training set*

Let $\{\Gamma_i \mid i = 1,2,3\ldots M\}$ be a training set Where $\Gamma$is a face Image ($N^2$).

*Step 2: Calculate mean image of all training samples.*

$$\Psi = \frac{1}{M}\sum_{i=1}^{M}\Gamma_i$$

*Step 3: Calculates the difference Images by subtracting the training set vector by the mean image. Let us call this matrix as the variation matrix.*

$$\Phi_i = \Gamma_i - \Psi$$

Where $A = [\Phi_1, \Phi_2, ..., \Phi_M]$ represents how each of original image varies from mean image.

*Step 4: Calculate Eigenvectors and Eigenvalues of covariance matrix $AA^T$.*

$$C = \frac{1}{M}\sum_{i=1}^{M}\Phi_i\Phi_i^{\,T} = AA^T$$

As the covariance matrix C has dimensions of $N^2 \times N^2$, I need to calculate $N^2$ eigenvectors. For images of a significant size this is a large computational task. I can solve for the $N^2$ dimensional eigenvectors in this case by first solving the eigenvectors of an M x M matrix i.e. $A^TA$.

$$(A^TA)\, V_i = \lambda_i V_i$$

$$A\,(A^TA)\, V_i = A\,(\lambda_i V_i)$$

$$(AA^T)\,(A\, V_i) = \lambda_i\,(AV_i)$$

Where $V_i$ and $\lambda_i$ are eigenvectors and eigenvalues of the smaller (M x M) $A^TA$ matrix respectively. The eigenvectors of the larger $AA^T$ matrix can be computed by calculating $AV_i$. The eigenvectors are sorted in descending order of eignevalues. They are shown in Fig. 4.

$$U_i = AV_i = [\Phi_{1,}, \Phi_2, \ldots, \Phi_M] \times \begin{bmatrix} v_1^i \\ \cdot \\ v_k^i \\ \cdot \\ v_M^i \end{bmatrix} = \sum_{k=1}^{M} v_k^i \Phi_k$$

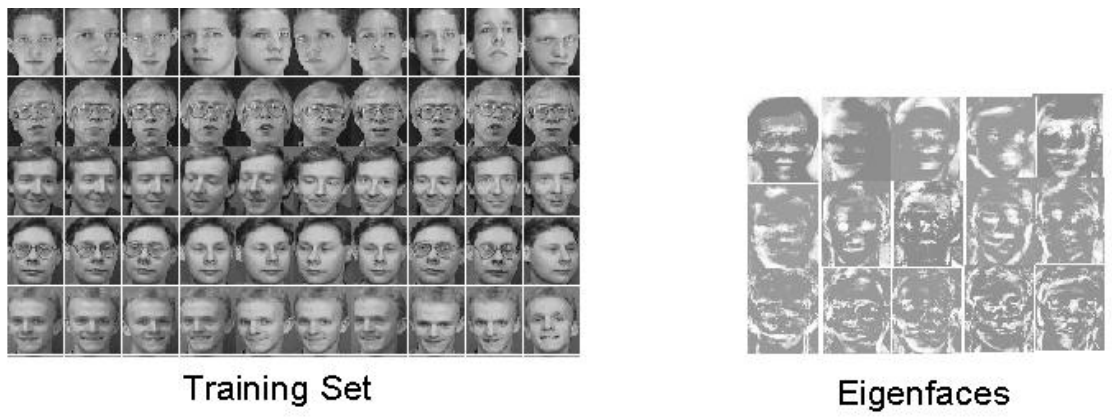

**Fig. 4:** Eigenfaces has a face like appearance

*Step 4: Represent face image using Eigenfaces*

$$W_i = \frac{U_K^T(\Gamma_i - \Psi)}{\lambda_K} \quad K=1,2\ldots M$$

where $W_i$ is a weight vector (i.e. $W_i$ (1,i) denotes percent that first eigeface represents image i , W(2, i) denotes percent that second eigeface represents image i, and so on)

To classify an input image following steps performed.

*Step 5: Convert test image into vector $\Gamma$*

*Step 6: Maps test image into Eigenfaces "face space".*

$W_k = U_k^T (\Gamma - \Psi), k = 1,\ldots, M'$

The weights form a feature vector,

$\Omega^T = [W_1 W_2 \ldots W_M]$

The feature vectors obtained from training set is used to train the neural network and feature vector of test image is used to simulate the neural network.

2.2.2. Independent Component Analysis

In the technique of ICA, one seeks to obtain completely independent components, which constitute complete faces. The basic idea is that any face image is a unique linear combination of these independent components.

$R=AU \Rightarrow U= W_I R$ where $W_I =A^{-1}$

Here, R = face images, A = unknown mixing matrix and U= statistically independent

It is important that these components should not only be de-correlated but completely independent from the point of view of higher order statistics as well.

I have used an algorithm proposed by Bell and Sejnowski[9] for separating the statistically independent components of a dataset.

The training face images are decomposed in statistically independent components known as a basic images as shown in Fig. 5. This information is used to encode the input face images and compare individuals as shown in Fig. 3b.

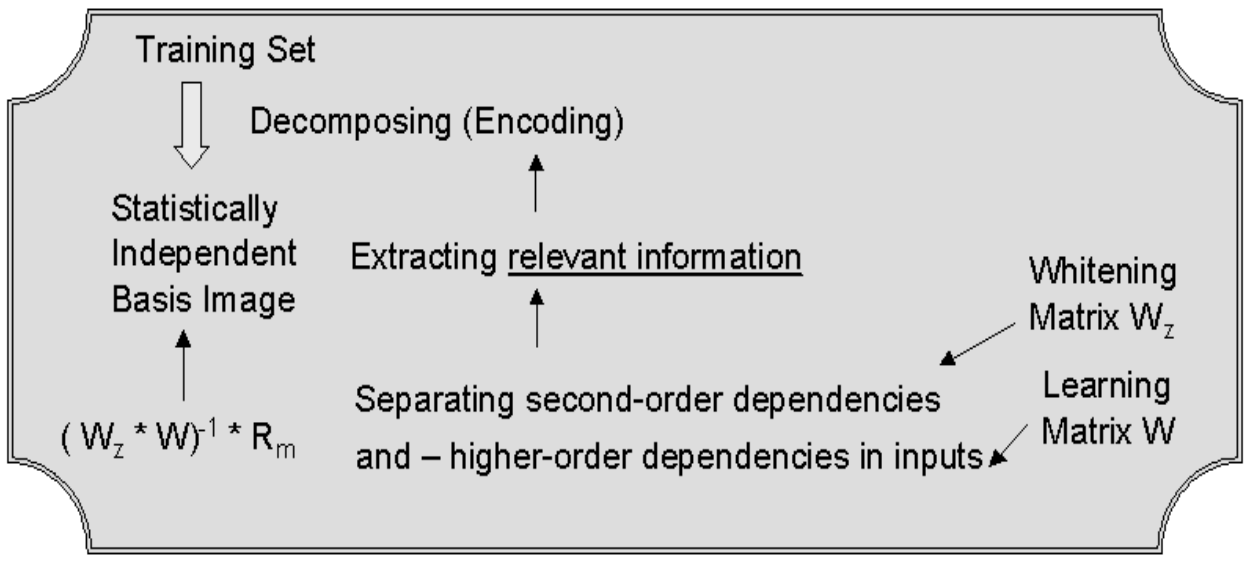

**Fig. 5:** Decomposing the training face images into statistically independent component.

***The algorithm is as follows:***

The ICA algorithm produced a matrix $W_I=W*W_Z$, Where $W_Z$ is a whitening matrix and W is a learning matrix. Here I am assuming that dimension reduction is already applied on training images either by LDA or PCA.

***Training Set***

*Step 1: Perform "Sphering" (step prior to learning) on training set*

*Centering:*

The row means are subtracted from the dataset, R, and then R is passed through the whitening filter

*Whitening:*

$W_Z = U \times (E)^{-1/2}$

where U and E are eigenvectors and eigenvalues respectively.

This step removes both the first and the second-order statistics of the data; both the mean and covariance are set to zero and the variances are equalize.

*Step 2: Calculate W iteratively*

In order to calculate W, I have sequence of pass through training data until old value of W and new value of W points in same direction refer to [9].

*Step 3: Calculate $W_I$*

$$W_I = W * W_Z$$

*Step 4: Calculate basis image (independent component). They are shown in Fig. 6.*

$$B = R * W_I^{-1}$$

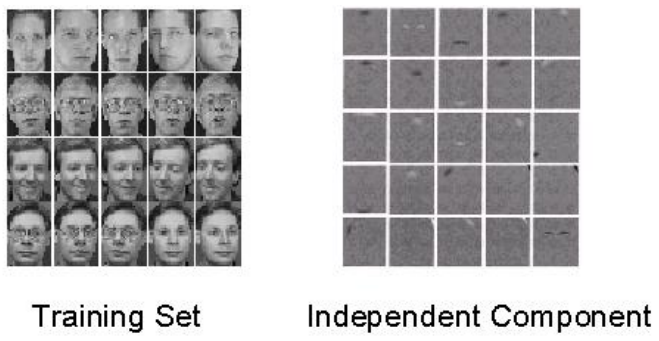

**Fig. 6:** Statistically independent component

To classify an input image following steps performed.

*Step 5: Project Test image into Eigenfaces (if dimension reduction applied prior to ICA)*

*Step 6: Compute coefficient $B_{test}$*

$$B_{test} = R_{test} * W_I^{-1}$$

Both B and $B_{test}$ are used as feature vector to train the neural network and simulate the neural network respectively.

## 2.3. Classification

Neural networks have been employed and compared to conventional classifiers for a number of classification problems. The results have shown that the accuracy of the neural network approaches equivalent to, or slightly better than, other methods. Also, due to the simplicity, generality and good learning ability of the neural networks, these types of classifiers are found to be more efficient[10]. The most popular neural network algorithm is back-propagation algorithm (a type of gradient decent method), I have used multi-layer feed-forward neural network on which backpropagation algorithm performs.

I have an input layer (i) consisting of input nodes and an output layer (k) consisting of output nodes. The input nodes are connected to the output nodes via one or more hidden layers (j). The nodes in the network are connected together, and each of the links has a weight associated with it. The output value from a node is a weighted sum of all the input values to the node. By changing the different weights of the input values I can adjust the influence from different input nodes. For face recognition the input nodes will typically correspond to image pixel values from the face image. The output layer will correspond to classes or individuals in the database. Each unit in the output layer can be trained to respond with +1 for a matching class and -1 for all others. In practice real outputs are not exactly +1 or -1, but vary in the range between these values. Classification is done by finding the output neuron with the maximal value. Then a threshold algorithm can be applied to reject or confirm the decision.

*Multi-layer feed-forward neural network based classifier design is explained as follows:*

The nodes in hidden layer and number of hidden layer are selected by trial and error; here I use one hidden layer with 70 neurons.

Step 1: Assemble training data (both input and output)

It takes input as Eigenfaces (U) (only first M' > M) and Basic Independent components (F) for PCA and ICA system respectively, and output as array (259 x 37) (259 face images and 37 feature considered)

Step 2: Create neural network and initialize it's parameter

Step 3: Train the neural network as mention in [10].

Step 4: Simulate the network response to an input image (s).

Output of both neural networks is given to combiner (fusion process).

### 2.4. Fusion Process

"Two heads are better than one" could be the basis premise of fusion.

If the score functions are directly comparable or if there exists at least an acceptable transformation scheme to make the involved classifiers comparable, score based strategies are good ways for decision process. In this paper, NN is used as a classifier for both systems, so naturally outputs of both systems are in same format hence select score based strategy as combiner [11].

The algorithm is as follows

```
Step 1: Assemble of output of both classifiers

Step 2: Set threshold values

Step 3:
If both classifiers classified an input image to same class label then
    If PCA score value > threshold (PCA) and ICA score values > threshold (ICA) then
        Classified an input image to class label (PCA / ICA) and return
    Else
        Denied access and return
    End
Else
    Goto step 4
End
Step 4:
If PCA score value > ICA score value then
    If  PCA score value  > threshold (PCA) then
        Classified an input image to class label (PCA) and return
    Else
        Denied access and return
    End
Else
    Goto step 5
End
Step 5:
If ICA score value  > threshold (ICA) then
    Classified an input image to class label (ICA)
    Return
Else
    Denied access and return
End
```

## 3. EXPERIMENTAL RESULT

The proposed system has been implemented using MATLAB.

### 3.1. Face Database

Two face databases are used in experiments, Olivetti research laboratory (ORL) face database and Shimon Edelman database.

The ORL database consists of 400 images acquired from 40 persons with variations in facial expression (e.g. open / close eyes, smiling / non-smiling), and facial details (e.g. wearing glasses / not wearing glasses). All images were taken under a dark background, and the subjects were in an upright frontal position, with tilting and rotation tolerance up to 20 degree. All images are gray scale with a 92*112 pixels resolution. Fig. 7 shows two individual samples in ORL database.

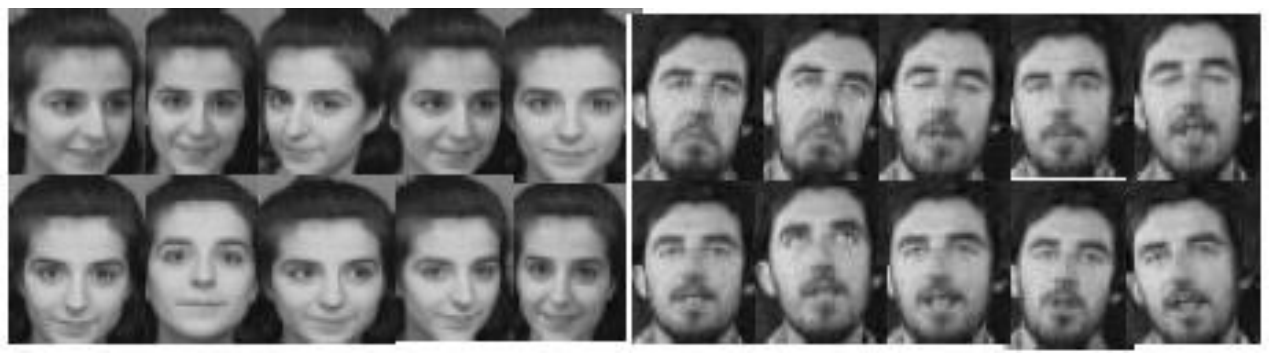

**Fig. 7:** Examples of two individual's face images in ORL database

There are 16 well-controlled images of each of 11 faces in the Shimon Edelman's database. All faces are of males without distinctive features such as glasses, beards, or mustaches. All images were taken by the same camera under tightly controlled conditions of illumination and viewpoint. The frontal view and natural expression face images under 16 different illumination directions are considered as shown in Fig. 8.

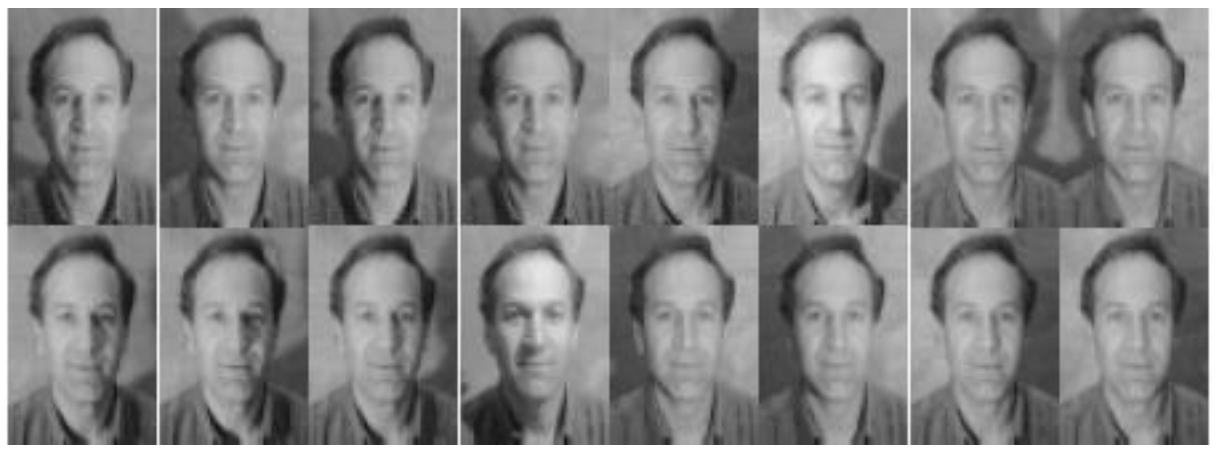

**Fig. 8:** An example of same face in Shimon face database

### 3.2. NN architecture and Parameters

In order to get better accuracy, it is important to clear out which NN training parameters were kept fixed and which varied during the upper tasks.

I have tested the system by changing different parameter of NN to get efficient architecture. I test system by varying learning rate {0.1,0.2,0.3,0.5,0.8} and momentum term {0,0.5,0.9} as shown in Fig. 9. It shows that, a very small learning rate and a normal momentum term will be advisable. I have also tested system for different number of neurons in hidden layer; best output is obtained with 70 neurons in hidden layer

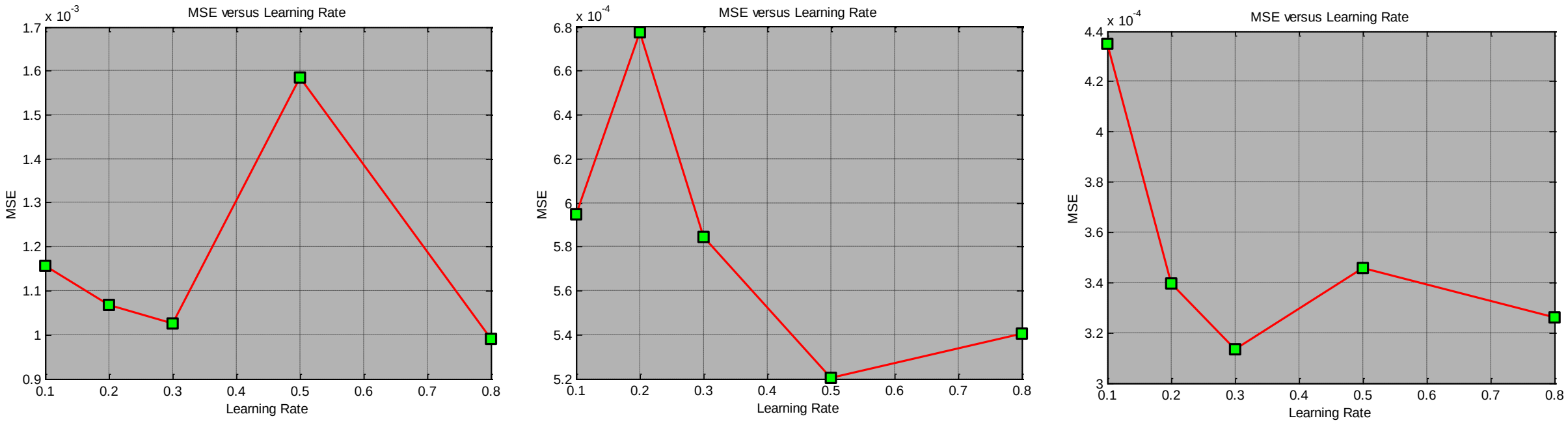

**Fig. 9 :** MSE versus learning rate **a)** when momentum term is 0; **b)** when momentum term is 0.5; **c)** when momentum term is 0.9

### 3.3. Comparison of PCA, ICA and Hybrid.

I divided each face database into two different sets: training set and testing set.

For ORL database, the training set and testing set are organized as follows.

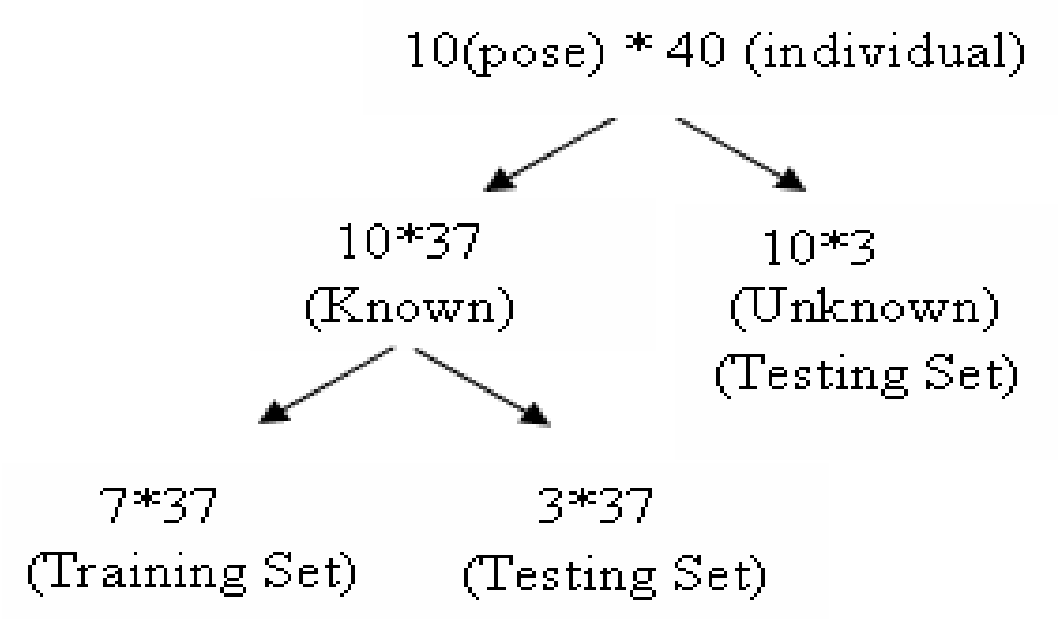

Out of 40 individuals, 37 individuals taken as known with their 7 poses for training and remaining 3 poses for testing. Other branch of tree shows that remaining 3 unknown individuals are taken as testing sets with all poses.

I have tested systems, with five different test sets. In test set 1, I considered remaining 3 pose of first known 20 individuals, which are rotated to only 5 to 10 degree. So, all the systems give 100% accuracy. In test set 2, I considered remaining 3 pose of known 17 individuals, which are rotated to 15 to 20 degree and some have open/close eye, smiling/not smiling. In test set 3, 4 and 5 I considered unknown person.

For Shimon Edelman's database, the training set and testing set are organized as follows.

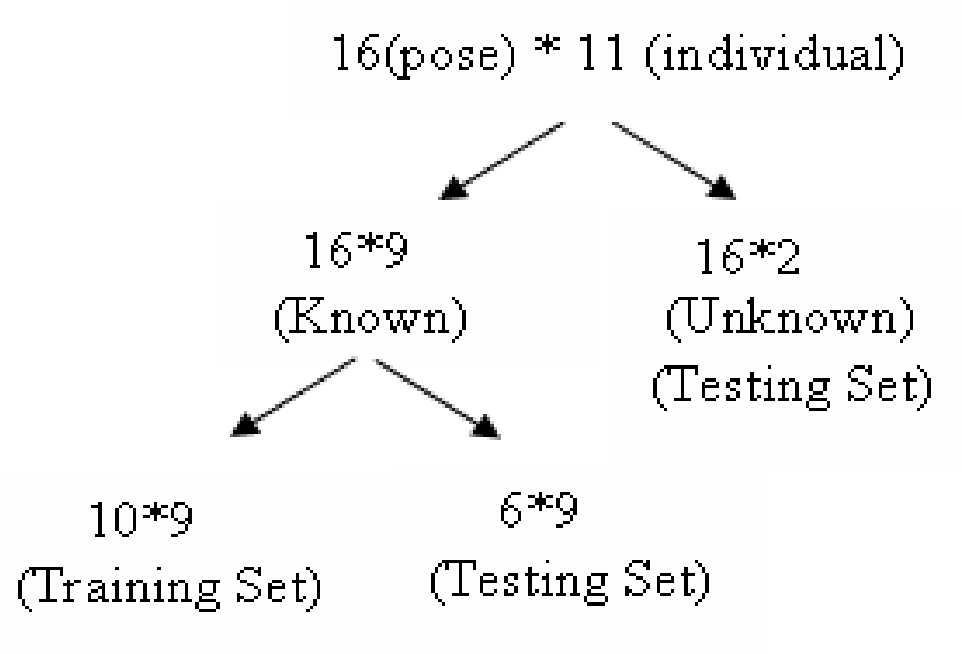

Out of 16 individuals, 9 individuals taken as known with their 10 poses for training and remaining 6 poses for testing. Other branch of tree shows that remaining 2 unknown individuals are taken as testing sets with all poses.

I have tested systems, with two different test sets. In test set 1, I considered remaining 6 pose of 9 individuals, and test set 2 contains all 16 poses of 3 unknown persons

Results of each of above test sets are shown in table 1 and table 2. It shows that for unknown person hybrid gives better result then ICA and PCA system. For known person, it gives good results compared to PCA and ICA system.

**Table 1:** Result on ORL face database

| Test Set | # of images | Accuracy (%) (PCA) | Accuracy (%) (ICA) | Accuracy (%) (HYBRID) |
|---|---|---|---|---|
| 1 | 60 ( 3 pose of known subject 1 - 20) | 100.00 (0 misclassified) | 100.00 (0 misclassified) | 100.00 (0 misclassified) |
| 2 | 51 ( 3 pose of known subject 21 - 37) | 94.12 (3 misclassified) | 96.07 (2 misclassified) | 98.04 (1 misclassified) |
| 3 | 9 ( 3 pose of unknown subject 38 - 40) | 33.33 (6 misclassified) | 44.44 (5 misclassified) | 98.04 (1 misclassified) |
| 4 | 9 ( 3 pose of unknown subject 38 – 40) | 11.11 (8 misclassified) | 66.67 (3 misclassified) | 98.04 (1 misclassified) |
| 5 | 12 ( 4 pose of unknown subject 38 – 40) | 8.33 (11 misclassified) | 50.00 (6 misclassified) | 83.33 (2 misclassified) |

**Table 2:** Result on Shimon Edelman's face database

| Test Set | # of images | Accuracy (%) (PCA) | Accuracy (%) (ICA) | Accuracy (%) (HYBRID) |
|---|---|---|---|---|
| 1 | 56 ( 3 pose of known subject 1 - 9) | 94.64 (3 misclassified) | 96.42 (2 misclassified) | 100.00 (0 misclassified) |
| 2 | 51 (16 pose of unknown subject 21 - 37) | 72.55 (15 misclassified) | 86.27 (7 misclassified) | 98.04 (1 misclassified) |

## 4. CONCLUSION

It shows that a plain PCA implementation is weaker than a plain ICA implementation since PCA is based on only second-order-statistics where as ICA is based on both second-order-statistics and higher-order-statistics. As "Two heads are better than one", *Hybrid Approach* gives better performance than both a plain PCA and ICA implementations.